\documentclass[runningheads]{llncs}

\usepackage[T1]{fontenc}
\usepackage{graphicx}
\usepackage{booktabs}
\usepackage{multirow}
\usepackage{array}
\usepackage{url}
\usepackage{amsmath}
\usepackage{amssymb}
\usepackage{microtype}
\usepackage{float}

\providecommand{\doi}[1]{\url{https://doi.org/#1}}

\begin{document}

\title{Scaling Hindi Quantum Natural Language Processing through Automatic Pregroup Supertagging}

\titlerunning{Hindi Pregroup Supertagging for QNLP}

\author{
Gautami Sanjay Naik\inst{1} \and
Krishna Bhatia\inst{2} \and
Mithun Paul Saint-Germain\inst{3} \and
H Aswath Babu\inst{1}
}

\authorrunning{Naik et al.}

\institute{
Indian Institute of Information Technology Dharwad, Dharwad, India\\
\email{gautaminaik2000@gmail.com}
\and
Fractal Analytics
\and
Arizona State University, Arizona, USA
}

\maketitle
\begin{abstract}
Quantum Natural Language Processing (QNLP) uses pregroup grammars to translate grammatical structure into diagrammatic representations and quantum circuits. Recent Hindi QNLP work has shown that Hindi-specific pregroup grammars can support grammar-sensitive compositional models, but grammatical type assignment is still largely manual, limiting scalability. This paper formulates automatic Hindi pregroup supertagging as a token-level classification task. Using a manually annotated corpus of 380 Hindi sentences, we evaluate lexical, contextual, prompting-based, lexical-repair, and suffix/morphology-aware methods. Results show that simple lexical and contextual models are strong in this low-resource setting: contextual backoff achieves the best completed accuracy of 64.56\%, while raw Qwen2.5 prompting reaches only 11.65\%. Lexical repair raises LLM-assisted prediction to 64.08\%, demonstrating the value of constraining generative outputs with symbolic grammar knowledge. Diagnostic analysis further shows that seen and unambiguous tokens are much easier than unseen tokens, and suffix/morphology features improve karaka-token accuracy but not overall performance. These results show that automatic Hindi pregroup assignment is feasible and can reduce reliance on manual annotation in future multilingual QNLP pipelines.

\keywords{Hindi NLP \and Quantum Natural Language Processing \and Pregroup Grammar \and Supertagging \and Large Language Models}
\end{abstract}

\section{Introduction}

Quantum Natural Language Processing (QNLP) combines compositional grammar, category theory, and quantum-inspired semantic representations. In the DisCoCat framework, grammatical reductions determine how word meanings compose into sentence meanings \cite{coecke2010mathematical}. Because pregroup grammars are compact closed categories, their reductions can be interpreted as string diagrams, tensor networks, and quantum circuits \cite{lambek1999type,lambek2008pregroup,coecke2020foundations,meichanetzidis2020qnlp,kartsaklis2021lambeq,lorenz2023qnlp}. Thus, grammatical type assignment is not only a preprocessing step; it directly determines the diagrammatic and circuit structure used by downstream QNLP systems.

Hindi is an important target language for QNLP because it is widely used and structurally different from English. It exhibits relatively flexible word order, rich morphology, postpositions, auxiliary constructions, and karaka-style dependency relations. These properties make direct transfer of English-style grammar resources insufficient and motivate Hindi-specific pregroup inventories. Recent work has begun extending QNLP beyond English and developing grammar-sensitive Hindi sentiment classification pipelines \cite{srivastava2023enabling,naik2026extending}. However, these systems still depend heavily on manually assigned pregroup types.

This paper addresses the bottleneck of automatic Hindi pregroup supertagging: assigning each Hindi token a valid pregroup type from a Hindi-specific grammar inventory. This task is related to supertagging in lexicalized grammars, where rich lexical categories encode much of the information needed for parsing \cite{bangalore1999supertagging,hockenmaier2007ccgbank,clark2007wide,yoshikawa2017astar,tian2020gcn}. While large language models have recently shown strong results on CCG and Lambek categorial grammar supertagging \cite{zhao2024llm}, it remains unclear whether raw LLM prompting can reliably generate exact symbolic pregroup types for Hindi.

We evaluate lexical majority, contextual prediction, contextual backoff, LLM prompting, lexical repair, and suffix/morphology-aware ranking. The study is motivated by three questions: whether Hindi pregroup types can be predicted automatically, whether LLM prompting improves over lexical-contextual baselines, and whether symbolic repair and morphology-aware features help in a low-resource Hindi setting. We also keep the work connected to modern Hindi and Indic NLP resources \cite{kunchukuttan2020indicnlp,chouhan2024hindillm}, transformer-based language modeling \cite{vaswani2017attention,brown2020language}, and the Qwen2.5 model family used for prompting \cite{yang2024qwen25}. Our main finding is that contextual lexical methods outperform raw prompting, and that LLM outputs become useful only when constrained by symbolic lexical repair.

\section{Methodology}

The research uses pregroup supertagging for Hindi as an automatic process at the level of tokens. Our goal is to identify the grammatical type (pregroup) that each token belongs to in a given Hindi sentence. As part of assessing whether it is possible to do this automatically, we performed experiments on a manually labeled corpus of Hindi pregroups and evaluated three different approaches (lexical, contextual, and LLM-based) within a single framework.

\subsection{Dataset and Task}

The experiments are performed on a manually annotated Hindi pregroup grammar corpus developed for Hindi Quantum Natural Language Processing. The corpus contains 380 sentence records. After alignment filtering, 376 sentence records and 2,048 tokens were used. The split contains 303 training sentences, 37 validation sentences, and 36 test sentences, with 206 test tokens. The task is Hindi pregroup supertagging, evaluated at the token level.

In the test set, 166 of 206 tokens are seen in the training split, while 40 tokens are unseen. Among the seen test tokens, 115 occur with ambiguous surface forms, meaning that the same lowercased token has more than one type in the training set. The test set also contains 12 karaka-bearing tokens, 71 tokens involving \texttt{n\_11}, and 16 tokens involving \texttt{n\_12}.

\subsection{Experimental Framework}

\begin{figure}[!t]
\centering
\includegraphics[width=0.68\textwidth]{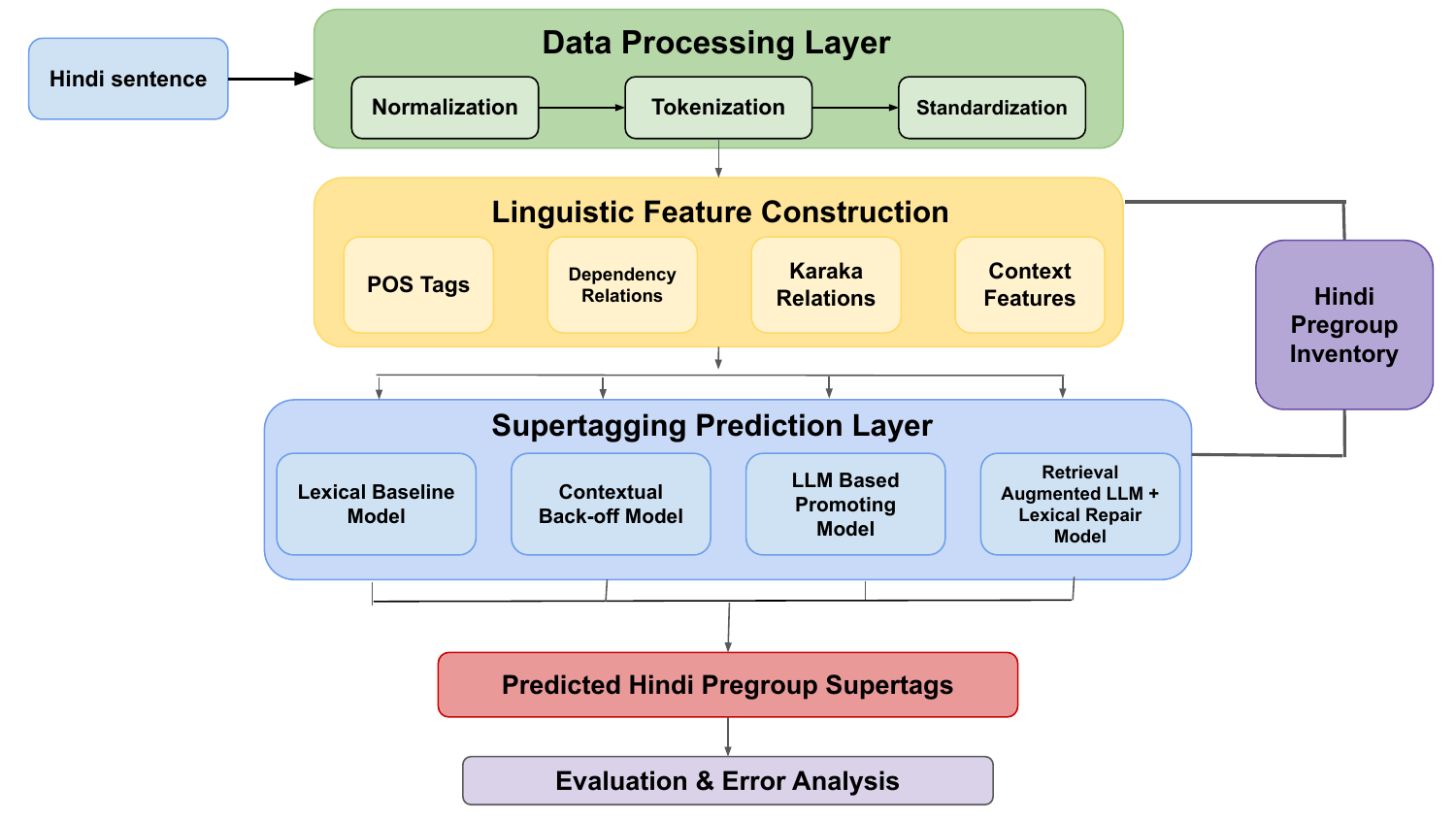}
\caption{Pipeline for automatic Hindi pregroup supertagging.}
\label{fig:pipeline}
\end{figure}

Automated Hindi pregroup supertagging was framed as a supervised prediction task. The complete workflow is shown in Figure~\ref{fig:pipeline}. The system begins with Hindi sentence preprocessing, constructs lexical and contextual features, predicts pregroup types using several models, and evaluates the predicted types against manually annotated gold labels.

\subsection{Methods Evaluated}

We evaluate global majority, lexical majority, local-context models using previous and next token information, contextual backoff, zero-shot and few-shot LLM prompting, repeated-input prompting, lexical repair, and suffix/morphology-aware ranking. The contextual backoff model uses the hierarchy
\[
(prev,w,next) \rightarrow (prev,w) \rightarrow (w,next) \rightarrow w \rightarrow global.
\]
LLM prompting is performed with Qwen2.5-1.5B-Instruct. The model is given a Hindi sentence and fixed tokenization and is instructed to output one pregroup type per token. Lexical repair validates generated labels against the training-set type inventory: for seen words, unsupported predictions are replaced by contextual backoff predictions; for unseen words, invalid or missing outputs are also repaired using contextual backoff. The suffix/morphology-aware candidate ranker uses weighted evidence from word identity, local context, position, suffixes, and global type counts.

\subsection{Evaluation Metrics and Implementation}

All models are tested using the same fixed train-validation-test split. The main metric is token-level exact-match accuracy. A prediction is counted as correct only when the predicted pregroup type exactly matches the manually annotated gold type. Additional evaluations are reported for seen and unseen words, ambiguous lexical items, karaka-containing grammatical categories, noun-class categories, oracle lexical coverage, and suffix/morphology behavior.

\section{Results and Discussion}

This Section will include presentation of experimental results based on evaluation of Hindi Pregroup Supertagging Approaches and the implications these have for automatic Grammatical Type Assignment. Results are displayed via table and graphically. Following the display of results, we provide a discussion of trends noted from the results and how they contribute to resolving the difficulty of manual Hindi Pre-group Annotation.

\subsection{Overall Performance of Supertagging Methods}

Table~\ref{tab:main-results} provides evidence regarding the total success rate of the examined algorithms in Hindi Pre Group Supertagger Task.

\begin{table}[!t]
\caption{Main results on the 36-sentence, 206-token test set.}
\label{tab:main-results}
\centering
\scriptsize
\setlength{\tabcolsep}{3pt}
\begin{tabular}{p{3.0cm}rrrp{2.8cm}}
\toprule
System & Correct / 206 & Exact acc. (\%) & Sent. exact (\%) & Notes \\
\midrule
Global majority & 33 & 16.02 & 0.00 & Label prior only \\
Lexical majority & 119 & 57.77 & 5.56 & Word identity \\
Context word-next & 127 & 61.65 & 11.11 & Right context \\
Context previous-word & 128 & 62.14 & 5.56 & Left context \\
Context previous+word+next & 130 & 63.11 & 8.33 & Full local context \\
\textbf{Contextual backoff} & \textbf{133} & \textbf{64.56} & \textbf{13.89} & Best completed system \\
Suffix/morph candidate ranker & 118 & 57.28 & 11.11 & Helps karaka, hurts overall \\
Best raw Qwen prompting & 24 & 11.65 & 0.00 & \(k=3\), repeat \(=1\) \\
Qwen + lexical repair & 132 & 64.08 & 13.89 & Repair-dominated \\
\bottomrule
\end{tabular}
\end{table}

As demonstrated by the data, there are considerable variations in terms of how successful each method was. Compared to other methods, the Global Majority baseline has shown the least amount of success; this indicates that designating one category for every item (i.e., a ``one size fits all'' or global majority) does not adequately reflect the grammar of Hindi. As illustrated by the Lexical Majority model, using word level lexical features resulted in an increase of nearly 30\% in the ability of a model to correctly predict pre groups.

An additional increase in the percentage of correct predictions of pregroups was evident when contextual information was used. Each of the three methods that utilize context (Previous Word, Word+Next, and Contextual Backoff), have been able to improve upon the lexical based model. Between these three methods, the highest token-level accuracy has been demonstrated by the Contextual Backoff method. This suggests that utilizing contextual information from the immediate area surrounding a word can help resolve many grammatical ambiguity issues.

In addition, it may be noted that although prompt-based LLMs appear to be vastly superior in terms of their breadth of knowledge with regard to linguistics, they were not capable of generating accurate symbolic pre group labels for items in the test set.

\subsection{Analysis of LLM-Based Prediction}

To evaluate the use of a Large Language Model (LLM) for Hindi pregroup supertagging, several approaches to prompting were investigated. Results from these experiments are shown in Table~\ref{tab:qwen-prompting} and Figure~\ref{fig:prompting-ablation}.

\begin{table}[!t]
\centering
\caption{Qwen2.5-1.5B raw prompting.}
\label{tab:qwen-prompting}
\scriptsize
\setlength{\tabcolsep}{2pt}
\begin{tabular}{lrrp{2.2cm}}
\toprule
Prompt & Rep. & Acc. & Interpretation \\
\midrule
\(k=0\) & 1 & 0.00 & Fails exact output \\
\(k=0\) & 3 & 0.00 & No zero-shot gain \\
\(k=3\) & 1 & 11.65 & Best raw result \\
\(k=3\) & 3 & 5.83 & Repetition hurts \\
\bottomrule
\end{tabular}
\end{table}

Prompt-based predictions alone did not yield satisfactory pregrouping for Hindi. Although few shot prompted systems performed better on average than those using zero shot prompting systems, their accuracies were still well below those obtained with contextualized systems. Repetition of the input sentence as part of the prompt was found to have an effect on accuracy at best marginal.

Thus it appears that while LLMs may be able to understand the syntactic properties of a Hindi sentence, generating explicit symbolic grammatical representations that are required by the pregroup framework does not appear to be possible. This finding suggests that the difficulty of the task is not only semantic or syntactic, but also representational: the model must generate exact symbolic type strings from a sparse and highly structured grammar inventory.

\subsection{Effect of Lexical Repair and Diagnostic Analysis}

Improving the predictive quality of LLMs through lexical repair techniques is also examined in this study. Results related to the analysis of these lexical repairs are shown in Table~\ref{tab:repair-diagnostic}.

\begin{table}[!t]
\caption{Lexical repair and diagnostic analysis.}
\label{tab:repair-diagnostic}
\centering
\scriptsize
\setlength{\tabcolsep}{4pt}
\begin{tabular}{lrrr}
\toprule
Bucket / System & Tokens & Correct & Accuracy (\%) \\
\midrule
Best raw Qwen prompting & 206 & 24 & 11.65 \\
Qwen + lexical repair & 206 & 132 & 64.08 \\
Contextual backoff & 206 & 133 & 64.56 \\
\midrule
Seen tokens & 166 & 122 & 73.49 \\
Unseen tokens & 40 & 11 & 27.50 \\
Ambiguous seen tokens & 115 & 77 & 66.96 \\
Unambiguous seen tokens & 51 & 45 & 88.24 \\
Karaka-bearing tokens & 12 & 6 & 50.00 \\
\texttt{n\_11} tokens & 71 & 42 & 59.15 \\
\texttt{n\_12} tokens & 16 & 6 & 37.50 \\
\bottomrule
\end{tabular}
\end{table}

Incorporating lexical repair into our models resulted in a substantial improvement in the quality of their output. Use of additional constraint on the model's output based on contextual data from its training set led to fewer instances of incorrect or ``invalid'' output by reducing the number of errors produced in relation to grammar. Additionally, it provided diagnostic evidence that the accuracy of predictions made by the model regarding previously encountered words (i.e., those included within the training vocabulary) far exceeded those for novel (unseen) word forms. Furthermore, as indicated by the results, there was an even greater difference in terms of accuracy for both ambiguous grammatical categories (i.e., those which have multiple possible classifications) and unambiguous lexical items (those with single classification).

The oracle word-candidate analysis further supports this interpretation. When the correct type was required to appear among the top lexical candidates for the same word, the top-1, top-3, and top-5 oracle accuracies were 52.43\%, 62.62\%, and 63.59\%, respectively. For seen words, these values increased to 65.06\%, 77.71\%, and 78.92\%, while unseen-word oracle accuracy remained 0.00\%. This shows that lexical memory provides a strong but incomplete ceiling: it is highly useful for words observed in training, but it cannot directly solve the unseen-token problem.

\subsection{Suffix/Morphology Analysis}

Additional analysis was performed to examine whether morphology-aware and suffix-based information could improve Hindi pregroup supertagging. Results are shown in Table~\ref{tab:suffix}.

\begin{table}[!t]
\caption{Suffix/morphology analysis.}
\label{tab:suffix}
\centering
\scriptsize
\setlength{\tabcolsep}{4pt}
\begin{tabular}{lrrr}
\toprule
System & Overall acc. (\%) & Unseen acc. (\%) & Karaka acc. (\%) \\
\midrule
Contextual backoff & 64.56 & 27.50 & 50.00 \\
Suffix/morph candidate ranker & 57.28 & 12.50 & 58.33 \\
\bottomrule
\end{tabular}
\end{table}

The suffix/morphology-aware candidate ranker achieved 57.28\% overall accuracy, which was below the contextual backoff model. However, it improved karaka-token accuracy from 50.00\% to 58.33\%. This suggests that suffix and surface-form features are linguistically informative for certain Hindi grammatical relations, especially karaka-linked categories, but they are not sufficient by themselves to outperform stronger lexical-contextual methods. At the same time, the lower unseen-token accuracy of the suffix/morphology ranker indicates that morphology-aware features should be integrated with contextual and lexical backoff strategies rather than used as a standalone replacement.

\subsection{Discussion}

The results provide significant insight into automatic Hindi pregroup supertagging. First, the stronger performance of contextual models shows that Hindi grammatical type assignment depends heavily on local lexical context. The contextual backoff model achieved the highest token-level accuracy of 64.56\%, indicating that a simple hierarchy of neighboring-word patterns can be effective in a low-resource Hindi pregroup setting.

Second, raw LLM prompting performed poorly for exact symbolic prediction. Although pregroup supertagging may appear similar to a text-generation task, it requires highly specific structural labels, and the best raw Qwen prompting condition reached only 11.65\% accuracy. This shows that unconstrained prompting alone is not reliable enough for Hindi pregroup supertagging.

Third, lexical repair substantially improved LLM-based prediction by constraining generated labels with symbolic grammar knowledge. The repaired LLM result of 64.08\% is close to contextual backoff, suggesting that most of the useful gain comes from lexical constraints rather than free-form generation.

Finally, the suffix and morphology experiment showed that morphology-sensitive features improved karaka-token accuracy from 50.00\% to 58.33\%, but did not improve overall or unseen-token accuracy. Thus, suffix-based information is useful for selected Hindi grammatical categories, but future work should integrate it with stronger lexical and contextual backoff mechanisms.

Overall, these results show that automatic Hindi pregroup assignment is feasible and can reduce reliance on manual annotation, providing a practical step toward scalable grammar-aware resources and multilingual QNLP pipelines.

\section{Conclusion}

This paper formulated automatic Hindi pregroup supertagging as a token-level classification task for scalable Hindi QNLP. Experiments on a manually annotated Hindi pregroup corpus showed that lexical and contextual methods are strong baselines, with contextual backoff achieving the best accuracy of 64.56\%. Raw LLM prompting performed poorly for exact symbolic type generation, while lexical repair improved LLM outputs by constraining them with training-set grammar information. Diagnostic results showed that unseen tokens remain the main bottleneck, indicating the need for larger annotated corpora, stronger morphology-aware models, and better grammar-aware constraints. Overall, automatic Hindi pregroup assignment is feasible and can reduce reliance on manual annotation in future multilingual QNLP pipelines.

\end{document}